\PassOptionsToPackage{unicode}{hyperref}
\PassOptionsToPackage{hyphens}{url}
\documentclass[
  11pt,
  letterpaper,
]{article}
\usepackage{xcolor}
\usepackage[margin=0.85in]{geometry}
\usepackage{amsmath,amssymb}
\usepackage{iftex}
\ifPDFTeX
  \usepackage[T1]{fontenc}
  \usepackage[utf8]{inputenc}
  \DeclareUnicodeCharacter{00D7}{\ensuremath{\times}}
  \DeclareUnicodeCharacter{03A9}{\ensuremath{\Omega}}
  \DeclareUnicodeCharacter{2192}{\ensuremath{\rightarrow}}
  \usepackage{textcomp} 
\else 
  \usepackage{unicode-math} 
  \defaultfontfeatures{Scale=MatchLowercase}
  \defaultfontfeatures[\rmfamily]{Ligatures=TeX,Scale=1}
\fi
\usepackage{lmodern}
\ifPDFTeX\else
\fi
\IfFileExists{upquote.sty}{\usepackage{upquote}}{}
\IfFileExists{microtype.sty}{
  \usepackage[]{microtype}
  \UseMicrotypeSet[protrusion]{basicmath} 
}{}
\usepackage{setspace}
\makeatletter
\@ifundefined{KOMAClassName}{
  \IfFileExists{parskip.sty}{%
    \usepackage{parskip}
  }{
    \setlength{\parindent}{0pt}
    \setlength{\parskip}{6pt plus 2pt minus 1pt}}
}{
  \KOMAoptions{parskip=half}}
\makeatother
\usepackage{longtable,booktabs,array}
\usepackage{calc} 
\usepackage{etoolbox}
\makeatletter
\patchcmd\longtable{\par}{\if@noskipsec\mbox{}\fi\par}{}{}
\makeatother
\IfFileExists{footnotehyper.sty}{\usepackage{footnotehyper}}{\usepackage{footnote}}
\makesavenoteenv{longtable}
\providecommand{\tightlist}{%
  \setlength{\itemsep}{0pt}\setlength{\parskip}{0pt}}
\usepackage{microtype}
\usepackage{tikz}
\usetikzlibrary{positioning,arrows.meta,shapes.geometric,fit}
\usepackage{bookmark}
\IfFileExists{xurl.sty}{\usepackage{xurl}}{} 
\hypersetup{
  pdftitle={Beyond Memory: A Templated Substrate for Heterogeneous Collaborative Knowledge Work with LLM Agents},
  pdfauthor={Priscila Saboia Moreira (CRC, University of Notre Dame); Christopher Sweet (CRC, University of Notre Dame)},
  hidelinks,
  pdfcreator={LaTeX via pandoc}}

\title{Beyond Memory: A Templated Substrate for Heterogeneous
Collaborative Knowledge Work with LLM Agents}
\author{Priscila Saboia Moreira (CRC, University of Notre
Dame) \and Christopher Sweet (CRC, University of Notre Dame)}
\date{2026-05-26}

\begin{document}
\maketitle

{
\setcounter{tocdepth}{2}
\tableofcontents
}
\setstretch{1.1}
\section{Abstract}\label{abstract}

Research projects, educational efforts, and adjacent knowledge work
accumulate findings, decisions, and reasoning that future collaborators
rarely recover. The parts most useful to that work, including dead ends
and walked-back claims, are routinely excluded from publications and
shared code; future researchers re-attempt the same failures because no
record survives. LLM coding agents are common participants but hold no
persistent memory across sessions, and retrieval-augmented generation
over raw sources does not compound. The llm-wiki pattern (Karpathy,
2026; tonbi, 2026) addresses this by inserting an LLM-maintained,
interlinked wiki between raw sources and the agent. We present
\texttt{llm-wiki-memory-template}, a reusable, agent-aware
instantiation, and argue it is a substrate for \emph{heterogeneous
collaborative knowledge work} along three axes (multi-human,
multi-AI-agent, multi-domain) with each axis supported by a distinct
architectural element of the template (§4). The wiki is append-only by
convention, which preserves \emph{what did not work} alongside what did,
addressing a negative-result loss problem that publications and
code-sharing structurally cannot solve. Three deployed case studies and
one design report cover the axes individually: a solo research lineage
that preserves abandoned iterations; a two-author project whose
retroactive audit revised two prior experiments' claimed 20-of-20
coverage down to 14 and 12 evidence-based answers, then to 18 and 18
after a fix, with the failure path preserved across the artifact; an
in-progress multi-agent deployment reported as a design; and a
cross-domain educational variant. We name failure-path preservation,
agent honesty, and appropriation as cross-cutting sociotechnical
properties of the artifact, not only of its technical mechanisms.

\subsection{1 Introduction}\label{introduction}

Research projects, educational efforts, and adjacent knowledge work
accumulate findings, decisions, and reasoning that are rarely captured
in a form a future collaborator, or a future version of the same person,
can recover. The dominant tools used to do the work are LLM coding
assistants: agents that read code, write code, draft documents, and
articulate ideas. These tools have transformed per-session productivity
but are amnesic across sessions: powerful, attentive collaborators with
no memory of what they helped produce yesterday.

Retrieval-augmented generation over raw sources is the standard response
and does not solve the problem. RAG retrieves chunks of the same raw
material on every query, and the material itself does not compound
between sessions. Worse, the parts of project history most useful to
future work, the dead ends, walked-back claims, and experiments whose
results turned out to be inflated under scrutiny, are routinely excluded
from publications and shared code entirely. Future researchers
re-attempt the same failures because no record of the earlier failure
survived.

The llm-wiki pattern (Karpathy, 2026; tonbi, 2026) addresses the
per-session memory problem by inserting an LLM-maintained, interlinked
wiki between raw sources and the agent: a small set of cross-referenced
Markdown pages with typed frontmatter, maintained by the agent itself
across sessions, and discoverable by humans through ordinary file
browsing or GitHub's wiki view. The contribution of this paper is
neither the pattern nor its implementation, but a \emph{reusable,
agent-aware, multi-axis collaborative template} that operationalizes the
pattern and demonstrates that it generalizes well past the
per-session-memory problem.

We present \texttt{llm-wiki-memory-template}, an instantiation that
ships with one-command bootstrap, parallel overlays for multiple coding
agents sharing one agent-agnostic core, a sync mechanism for derived
projects, attribution on every action, and a small set of discipline
gates and a verification gate that catch agents filing projections as
measured facts. The extension is the \emph{three-axis architectural
framing} (§4): the template is a substrate for \emph{heterogeneous
collaborative knowledge work} along three axes, each supported by a
distinct architectural element, with combinations handled by design.

The paper makes four contributions:

\begin{enumerate}
\def\labelenumi{\arabic{enumi}.}
\tightlist
\item
  \textbf{Failure-path preservation} as a sociotechnical property of
  LLM-mediated knowledge artifacts (§6.1). The wiki's append-only
  convention preserves abandoned approaches, walked-back claims, and
  audited-down numbers alongside the eventual best result. We argue this
  is the contribution most deserving of attention beyond research
  software engineering.
\item
  \textbf{A three-axis design framing} (§4): each axis mapped to a
  specific template mechanism. The composition claim is a \emph{design
  prediction} exercised one and two axes at a time; the
  three-axes-simultaneous case is a near-term opportunity.
\item
  \textbf{A reusable, agent-aware template instantiation} (§3) supplying
  discipline gates, verification gates, attribution logging, and a
  three-tier governance model.
\item
  \textbf{Three deployed case studies plus one design report} (§5)
  covering the axes individually.
\end{enumerate}

The methodology is interpretive trace analysis: for each case we read
the wiki's content and the git log of the wiki repository, treating
commits, page amendments, and cross-references as the observable
artifacts of articulation work. Numbers in §5 and §6 come from public
git logs (Cases A and D) or from saved per-experiment summaries (Case B,
private). The rest proceeds: §2 places the work in CSCW and adjacent
literatures; §3 describes the reusable core; §4 lays out the
architectural argument; §5 presents the cases; §6 synthesizes the
findings; §7 names limitations; §8 concludes.

\subsection{2 Background and related
work}\label{background-and-related-work}

We position the contribution against two literatures: CSCW (which
supplies the vocabulary the three-axis argument relies on) and recent
technical work on LLM memory (which supplies the comparison space for
the per-session problem). We also name the llm-wiki lineage and three
public extensions that appeared concurrently with our work.

\subsubsection{2.1 CSCW: common information spaces, articulation,
awareness,
appropriation}\label{cscw-common-information-spaces-articulation-awareness-appropriation}

The wiki is a \emph{common information space} (Schmidt and Bannon, 1992;
Bannon and Bødker, 1997): a shared artifact that supports cooperative
work asynchronously. The work of making knowledge legible within such a
space is \emph{articulation work} (Strauss, 1985, 1988); the pattern
operationalizes it as a first-class output, since typed-edge
frontmatter, cross-references, and per-action attribution are artifacts
of articulation made durable. The agent-overlay system (§4.2) is
\emph{boundary infrastructure} (Star and Bowker, 1999): a shared
backbone accessed through local interfaces tailored to each
participant's tooling. The variant tier (§4.3) is \emph{appropriation}
(Dourish, 2001): stable enough to recognize, open enough to extend, with
\emph{communities of practice} (Lave and Wenger, 1991) supplying the
framing for cross-domain variants. The
\texttt{by:\ \textless{}human\textgreater{}\ via\ \textless{}agent\textgreater{}}
attribution line (§3.3) treats provenance as an \emph{awareness}
primitive (Dourish and Bellotti, 1992) operating \emph{across sessions}
rather than \emph{across desks}.

\subsubsection{2.2 LLM memory and retrieval-augmented
generation}\label{llm-memory-and-retrieval-augmented-generation}

Retrieval-augmented generation (Lewis et al., 2020) indexes documents at
write time and retrieves at query time, addressing the
\emph{long-context} problem but not \emph{compounding}: nothing is
compiled, cross-referenced, or kept current between sessions.
Graph-structured indices (Edge et al., 2024; Gutiérrez et al., 2024,
2025) improve the retrieval substrate's structure but retain the
per-query rediscovery dynamic. MemGPT (Packer et al., 2023) treats
memory as a \emph{runtime} property of the agent. These mechanisms
manage an agent's \emph{internal} representation; the llm-wiki pattern
manages an \emph{external} artifact that humans and agents both read and
write, durable across sessions, users, and agents. The two layers
compose: an agent with strong internal memory still benefits from an
external wiki when collaborators need to read, audit, or extend it.

\subsubsection{2.3 The llm-wiki lineage and concurrent
extensions}\label{the-llm-wiki-lineage-and-concurrent-extensions}

The pattern was proposed by Karpathy (April 2026) and instantiated in a
working template repository by tonbi (April 2026). Three public
extensions appeared concurrently with our work. \emph{Beyond the Wiki:
Scaling Karpathy's LLM Wiki Pattern for Multi-Agent Production}
(redmizt, 2026) adds identity, security, concurrency, and a
knowledge-graph layer, and names the same agent-honesty failure mode we
target in §6.2. \emph{LLM Wiki v2} (rohitg00, 2026) adds confidence
scoring and explicit \emph{supersession} of stale claims by
higher-confidence successor pages. A third public rebuild (Ghelbur,
2026) adds scheduled agents, automatic synthesis, and AI-first note
structure. These extensions and ours converge on the same problem but
diverge on the response. The most consequential divergence is
\emph{supersession versus preservation}: \emph{LLM Wiki v2} overwrites
stale claims; our failure-path-preservation property (§6.1) retains them
in place. The broader argument that publication norms discard negative
results is rehearsed in metascience (Ioannidis, 2005); the pattern
pushes the discipline \emph{upstream} of publication into the artifact,
a different point of intervention.

\subsection{3 The template's reusable
core}\label{the-templates-reusable-core}

This section describes the parts of \texttt{llm-wiki-memory-template}
(CRC Research, n.d.) the rest of the paper relies on. Full source and
documentation live at
https://github.com/crcresearch/llm-wiki-memory-template (and in that
repository's own wiki, an instance of the pattern).

\subsubsection{3.1 Bootstrap and the wiki}\label{bootstrap-and-the-wiki}

A one-command bootstrap script (\texttt{scripts/instantiate.sh})
substitutes project-specific placeholders, scaffolds a wiki
sub-repository at \texttt{wiki/\textless{}repo\textgreater{}.wiki/} with
home, index, log, and schema pages, installs the chosen agent overlay
(\texttt{-\/-agent=\textless{}claude-code\textbar{}cursor\textbar{}none\textbar{}all\textgreater{}}),
optionally selects GitHub's wiki backend in place of the default
local-only sub-repo (\texttt{-\/-github-wiki}), and self-deletes. Each
wiki is a small set of cross-linked Markdown pages with YAML frontmatter
declaring page \emph{type}, parent (\texttt{up:}), and optional typed
edges (\texttt{extends:}, \texttt{supports:}, \texttt{criticizes:},
\texttt{source:}, \texttt{related:}). Body cross-references use
\texttt{{[}Display{]}(Page-Name)}; frontmatter edges use
\texttt{{[}{[}Page-Name{]}{]}} wikilinks. Three operations are codified:
\textbf{ingest} files or updates pages with bidirectional
cross-reference repair, an index update, and an appended log entry,
exposed as two type-specific commands (\texttt{wiki-experiment} for
experimental results, \texttt{wiki-source} for external source
documents); \textbf{query} reads the index, then the relevant pages, and
synthesizes (reusable answers filed as new pages), invoked as an agent
prompt rather than a dedicated command; \textbf{lint} has its own
command and runs a periodic health-check for orphans, dead links, stale
claims, missing frontmatter, and \texttt{untyped} pages. The wiki is
browsable via the file system, GitHub's native wiki view, or
Obsidian-style clients; the underlying
Markdown-plus-frontmatter-plus-git substrate is the same across
surfaces, which lowers the adoption bar for collaborators who do not run
a git client.

\subsubsection{3.2 Agent overlays and
sync}\label{agent-overlays-and-sync}

Each overlay's source lives in
\texttt{wiki/agents/\textless{}agent\textgreater{}/} with its own
\texttt{setup.sh} and \texttt{templates/}; the runtime artifacts emit to
agent-recognized locations (\texttt{.claude/} for Claude Code,
\texttt{.cursor/} for Cursor), separating the versioned source layer
from the active runtime layer and letting the agent-agnostic policy
files at \texttt{wiki/agents/} be referenced by every overlay rather
than copied. The Claude Code overlay ships slash commands and skills
(\texttt{.claude/commands/wiki-*.md},
\texttt{.claude/skills/wiki-*.md}), a wiki-scoped pre-approval allowlist
for
\texttt{git\ -C\ wiki/\textless{}repo\textgreater{}.wiki/\ add\ \textbar{}\ commit\ \textbar{}\ status\ \textbar{}\ log\ \textbar{}\ diff}
(never \texttt{push}), and an optional SessionStart hook that reminds
the agent the wiki is project memory and prints the available commands
at every session start; an optional \texttt{-\/-seed-memory} flag
installs a starting memory page so new projects begin with a small seed
rather than a cold start. The Cursor overlay ships
\texttt{.cursor/rules/wiki-*.mdc} with an always-apply memory rule and
three \emph{Agent Requested} operation rules. A ``none'' overlay ships
no agent files; the operations live in \texttt{CLAUDE.md} and
\texttt{SCHEMA\_\textless{}repo\textgreater{}.md} for the agent to read
directly. Derived projects pull base-template improvements via
\texttt{scripts/update-from-template.sh}; an \texttt{ALWAYS\_FILES}
allowlist controls what propagates (wiki init script, per-overlay setup,
agent-agnostic policy files, per-overlay surfaces, knowledge-graph
pipeline). Files not on the allowlist are project-owned and never
overwritten.

\subsubsection{3.3 Agent-honesty
mechanisms}\label{agent-honesty-mechanisms}

Three mechanisms apply across all overlays. \emph{Discipline gates}
(\texttt{wiki/agents/discipline-gates.md}) name \emph{universally-wrong
rationalizations} paired with the discipline that defeats them,
organized into three gate categories (design, verification, sequential)
and a skill-dependency chain that says which gates a given operation
must pass before commit. The \emph{verification gate}
(\texttt{wiki/agents/verification-gate.md}) is a pre-commit checklist
organized in four categories: \emph{numerical claims} (backed by real
script output rather than estimates, tagged with corpus and scope, no
cross-corpus gaps presented as direct), \emph{cross-references and
structure} (reciprocal typed-edges, body links resolving to existing
pages, valid frontmatter with required fields), \emph{index and log}
(correct-category index entry, substantive log bullets), and
\emph{honest reporting} (negative results filed truthfully). The
numerical-claims category is the explicit mechanism against the
projection-as-fact failure mode Case B's audit cascade caught (§5.2,
§6.2). \emph{Log-entry attribution} requires every entry in
\texttt{log\_\textless{}repo\textgreater{}.md} to carry a
\texttt{by:\ \textless{}human\textgreater{}\ via\ \textless{}agent\textgreater{}}
line as its first bullet (the human read from
\texttt{git\ config\ user.name}, not invented); each log append is
committed on its own, so \texttt{git\ blame} on the log is a faithful
per-entry record. Both policy files are agent-agnostic: every overlay
references them rather than reimplementing them, so policy stays
consistent as overlays proliferate.

\subsection{4 Three-axis architecture}\label{three-axis-architecture}

We claim the llm-wiki template is a substrate for collaborative
knowledge work along \emph{three} axes: collaboration \emph{across
humans}, collaboration \emph{across AI agents}, and reuse \emph{across
domains}. Each axis is supported by a distinct architectural element of
the template, and the three elements compose: a deployment can sit on
any one axis, any two, or all three at once without changing the
underlying pattern. This section makes the mapping precise (one
subsection per axis) and then describes how composition works in
practice (§4.4). The empirical demonstration of each axis is deferred to
the case studies in §5; §4 is the \emph{designed}-side argument.

Figure 1 summarizes the mapping. Each row is an axis; each row lists the
failure mode that would make collaboration on that axis impossible, the
architectural mechanism the template uses to forestall the failure, and
the section of this paper that demonstrates the mechanism on real
adoption.

\begin{figure}[h]
\centering
\begin{tikzpicture}[
  axislabel/.style={font=\bfseries\small, align=center, text width=2.2cm},
  header/.style={font=\bfseries\small, align=center},
  cell/.style={draw, rectangle, rounded corners=2pt,
               minimum width=3.6cm, minimum height=1.4cm,
               align=center, text width=3.4cm, font=\footnotesize,
               inner sep=4pt},
  arrow/.style={-{Latex[length=2mm]}, thick, gray!60}
]

\node[header] at (0, 2.6) {Failure mode};
\node[header] at (4.2, 2.6) {Architectural mechanism};
\node[header] at (8.4, 2.6) {Demonstrated in};

\node[axislabel] at (-3.6, 1.4) {Multi-\\human};
\node[cell] (r1c1) at (0, 1.4) {Inability to attribute claims to authors};
\node[cell] (r1c2) at (4.2, 1.4) {Wiki as common information space; \texttt{by:} line; append-only convention};
\node[cell] (r1c3) at (8.4, 1.4) {\S 5.1, \S 5.2};

\node[axislabel] at (-3.6, -0.2) {Multi-AI-\\agent};
\node[cell] (r2c1) at (0, -0.2) {Agent overwrite; policy drift across agents};
\node[cell] (r2c2) at (4.2, -0.2) {Agent-overlay layer; agent-agnostic policy files; one commit per log entry};
\node[cell] (r2c3) at (8.4, -0.2) {\S 5.3};

\node[axislabel] at (-3.6, -1.8) {Multi-\\domain};
\node[cell] (r3c1) at (0, -1.8) {Base bloat; variant drift across variants};
\node[cell] (r3c2) at (4.2, -1.8) {Three-tier governance; placement rule; \texttt{ALWAYS\_FILES}; graduation trigger};
\node[cell] (r3c3) at (8.4, -1.8) {\S 5.4};

\draw[arrow] (r1c1) -- (r1c2);
\draw[arrow] (r1c2) -- (r1c3);
\draw[arrow] (r2c1) -- (r2c2);
\draw[arrow] (r2c2) -- (r2c3);
\draw[arrow] (r3c1) -- (r3c2);
\draw[arrow] (r3c2) -- (r3c3);

\end{tikzpicture}
\caption{Three-axis architectural mapping. Each row is one collaboration axis (multi-human, multi-AI-agent, multi-domain). Within each row, the architectural mechanism (middle column) forestalls the failure mode (left) that would otherwise make collaboration on that axis impossible; the case study where the mechanism is demonstrated on real adoption is named on the right. The three axes are independent: the architecture supports any one, any two, or in principle all three at once.}
\label{fig:three-axis}
\end{figure}
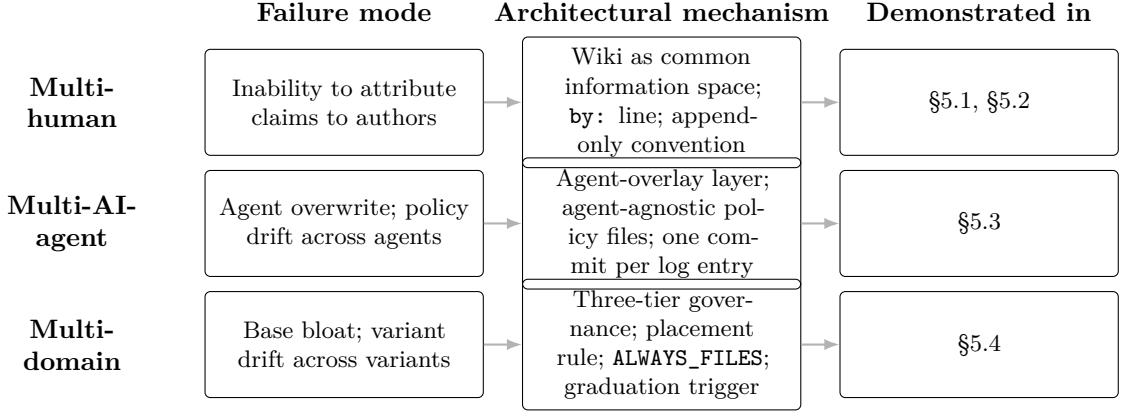

\subsubsection{4.1 Multi-human: wiki as common information
space}\label{multi-human-wiki-as-common-information-space}

The wiki itself is the substrate for multi-human collaboration. We treat
it as a \emph{common information space} (Schmidt and Bannon, 1992;
Bannon and Bødker, 1997): a shared artifact that supports cooperative
articulation work across participants who are not always present
together, without requiring synchronous coordination. The artifact is
the place where claims, findings, and decisions become legible to
collaborators who arrive later, and where the work of making them
legible (what Strauss calls articulation work) is itself preserved as
part of the project record.

Three template features support the multi-human axis specifically.
First, the wiki is \emph{append-only} by convention: pages are added and
amended, never deleted, so a researcher arriving at a page sees the
trajectory of claims rather than only the current snapshot. Second, the
typed-edge frontmatter described in §3 makes articulation work
\emph{explicit}: when one researcher writes a \texttt{supports:} edge,
that edge is an instruction to a future reader (or agent) about how to
combine the two pages, not a vague ``see also.'' Third, the \texttt{by:}
line on every log entry establishes per-action attribution: a future
audit of ``who claimed what when'' reduces to reading the log file. The
failure mode that would make multi-human collaboration impossible is the
inability to attribute claims to authors; the \texttt{by:} line
addresses that failure by construction.

What this axis does \emph{not} require is real-time presence, shared
editing sessions, or a centralized coordinator. The wiki's load-bearing
property is that it supports asynchronous, attributed articulation work,
which is enough for two-author projects (§5.2) and scales to small
teams.

\subsubsection{4.2 Multi-AI-agent: agent overlays as
tool-pluralism}\label{multi-ai-agent-agent-overlays-as-tool-pluralism}

The agent overlay system is the substrate for multi-AI-agent
collaboration. Different coding assistants (Claude Code, Cursor, future
others) read and write the same wiki, but each reads it through its own
\emph{overlay}: a per-agent set of slash commands, skills, rules, or
settings files (§3) that translates the agent-agnostic operations of the
template (ingest, query, lint, verify) into the surfaces that agent
supports. The overlay system is a form of \emph{boundary infrastructure}
in the sense of Star and Bowker: different participants access one
shared artifact through local interfaces tailored to each participant's
tooling, while the underlying artifact retains a single representation.

Two template features support the multi-AI-agent axis specifically.
First, two policy files (\texttt{discipline-gates.md} and
\texttt{verification-gate.md}) are \emph{agent-agnostic}: every overlay
references them rather than reimplementing them, so policy stays
consistent across overlays even as overlays proliferate. Without this
DRY discipline, two overlays would inevitably drift into incompatible
interpretations of the same operation. Second, the \texttt{by:} line's
\texttt{via\ \textless{}agent\textgreater{}} half makes the agent
visible alongside the human. A future audit of a multi-agent deployment
can distinguish which claims came in via Claude Code, which via Cursor,
and so on, without requiring the human to remember.

The failure mode that would make multi-AI-agent collaboration impossible
is \emph{agent overwrite}, where one agent's writes invalidate
another's. The template forestalls this with two conventions: the
one-commit-per-log-entry rule (§3) preserves per-action provenance even
when multiple agents are active, and the verification gate (§3) prevents
an agent's writes from landing without satisfying basic consistency
checks.

\subsubsection{4.3 Multi-domain: three-tier governance as appropriation
infrastructure}\label{multi-domain-three-tier-governance-as-appropriation-infrastructure}

The three-tier governance model (base / variant / derived) is the
substrate for cross-domain reuse. The pattern began as a
research-software memory substrate but has been \emph{appropriated} by
educational, training, and workforce-development projects. Appropriation
in the sense of Dourish requires the artifact to remain stable enough
that distant adopters can recognize it, while being open enough that
adopters can extend it for their own purposes. The template's three
tiers preserve both properties. Figure 2 sketches the structure.

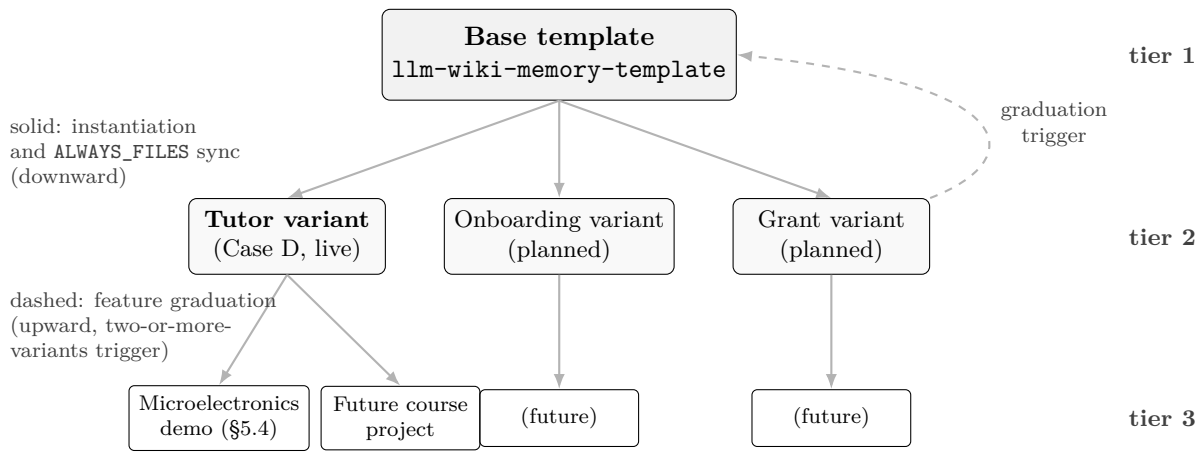
\begin{figure}[h]
\centering
\begin{tikzpicture}[
  base/.style={draw, rectangle, rounded corners=3pt,
               minimum width=4.4cm, minimum height=1.2cm,
               align=center, font=\small, fill=gray!10},
  variant/.style={draw, rectangle, rounded corners=3pt,
                  minimum width=2.6cm, minimum height=1cm,
                  align=center, font=\footnotesize, fill=gray!5},
  derived/.style={draw, rectangle, rounded corners=2pt,
                  minimum width=2.1cm, minimum height=0.75cm,
                  align=center, font=\scriptsize},
  flow/.style={-{Latex[length=2mm]}, thick, gray!60},
  grad/.style={-{Latex[length=2mm]}, thick, dashed, gray!60},
  tierlabel/.style={font=\footnotesize\bfseries, anchor=west, gray!50!black}
]

\node[base] (base) at (0, 3) {\textbf{Base template}\\\texttt{llm-wiki-memory-template}};

\node[variant] (v-tutor) at (-3.6, 0.6) {\textbf{Tutor variant}\\(Case D, live)};
\node[variant] (v-onb)   at (0, 0.6) {Onboarding variant\\(planned)};
\node[variant] (v-grant) at (3.6, 0.6) {Grant variant\\(planned)};

\node[derived] (d-mic)  at (-4.5, -1.8) {Microelectronics\\demo (§5.4)};
\node[derived] (d-fut1) at (-2.1, -1.8) {Future course\\project};

\node[derived] (d-fut2) at (0,   -1.8) {(future)};
\node[derived] (d-fut3) at (3.6, -1.8) {(future)};

\draw[flow] (base.south) -- (v-tutor.north);
\draw[flow] (base.south) -- (v-onb.north);
\draw[flow] (base.south) -- (v-grant.north);

\draw[flow] (v-tutor.south) -- (d-mic.north);
\draw[flow] (v-tutor.south) -- (d-fut1.north);
\draw[flow] (v-onb.south)   -- (d-fut2.north);
\draw[flow] (v-grant.south) -- (d-fut3.north);

\draw[grad] (v-grant.north east) .. controls (6.2, 1.6) and (6.2, 2.6) .. (base.east) 
  node[pos=0.5, right=4pt, font=\scriptsize, gray!50!black, align=center] {graduation\\trigger};

\node[tierlabel] at (7.4, 3)    {tier 1};
\node[tierlabel] at (7.4, 0.6)  {tier 2};
\node[tierlabel] at (7.4, -1.8) {tier 3};

\node[font=\scriptsize, gray!50!black, align=left, anchor=west] at (-7.4, 1.7) 
  {solid: instantiation\\and \texttt{ALWAYS\_FILES} sync\\(downward)};
\node[font=\scriptsize, gray!50!black, align=left, anchor=west] at (-7.4, -0.6) 
  {dashed: feature graduation\\(upward, two-or-more-\\variants trigger)};

\end{tikzpicture}
\caption{Three-tier governance model. The base template (tier 1) supplies generic llm-wiki scaffolding usable by any project. Variant templates (tier 2) are forks of the base that supply behavior contracts for *classes* of derived projects; the tutor variant of Case D is the first live instance, onboarding and grant variants are under design discussion. Derived projects (tier 3) are per-instance instantiations of either the base directly or a variant. Solid arrows show instantiation and the \texttt{ALWAYS\_FILES} sync contract (downward). The dashed arrow shows the graduation trigger: when two or more variants converge on a generic feature, the placement rule promotes it back to the base, and the variants delete it on the next rebase.}
\label{fig:governance}
\end{figure}

Three template features support the multi-domain axis specifically.
First, the \emph{placement rule} (Governance, §3) gives each new feature
an unambiguous home: a feature lives at the most-general tier that
benefits from it. The rule prevents the base template from accumulating
variant-specific features (``would a project that has nothing to do with
this variant \emph{still} benefit?''). Second, the
\texttt{ALWAYS\_FILES} sync contract (§3) defines exactly which files
propagate from base to variants and from variants to derived projects,
so improvements to the base reach all instantiations through a single
mechanism without manual re-application. Third, the rebase discipline
and the graduation trigger (Governance) give variants a procedure for
staying current and for promoting variant-specific features upstream
when two or more variants converge on them.

The failure modes that would make multi-domain reuse impossible are
\emph{base bloat} (variant-specific features leak into the base, making
it hostile to projects that don't need them) and \emph{variant drift}
(variants reimplement the same general feature in incompatible ways
because neither pushes upstream). The placement rule addresses the
first; the graduation trigger and the rebase discipline address the
second.

\subsubsection{4.4 Composition: combinations by
design}\label{composition-combinations-by-design}

The three architectural elements compose because they operate at
\emph{different layers} of the template and do not constrain one
another. The wiki layer supports multi-human collaboration regardless of
which overlays are in use; the overlay layer supports multi-agent
collaboration regardless of which domain variant the template was
instantiated from; the governance layer supports multi-domain reuse
regardless of how many humans or agents are active in any one
instantiation. A deployment can be characterized by which axes are ``lit
up'', and the architecture handles each combination by the union of the
relevant mechanisms.

Three combinations occur in our case studies. The two-author
web-scraping project (§5.2) is \emph{multi-human × single-domain ×
single-agent}: the wiki is the shared CIS, the \texttt{by:} line
attributes claims to the two researchers, the overlay is Claude Code
throughout. The in-progress multi-agent deployment of §5.3 is
\emph{multi-human × single-domain × multi-agent}: the overlay system is
the load-bearing piece, the \texttt{by:\ ...\ via\ ...} line
distinguishes humans-by-agent, and a single research computing center's
wiki is the shared CIS. The educational variant (§5.4) is
\emph{multi-human × multi-domain × single-agent}: the variant template
fork is the load-bearing piece, the wiki carries domain-specific content
(an Arduino curriculum) without bloating the base, and the same Claude
Code overlay handles ingest. No single case study uses all three axes
simultaneously yet; the architecture predicts that case would work, and
we identify it as a near-term opportunity in §8.

The three-axis architecture is a design framing; the composition claim
above is a \emph{design prediction}, supported by case studies that each
light up one or two axes but not yet by a case that lights up all three
simultaneously (§7). The paper's load-bearing empirical claim is the
failure-path-preservation property (§6.1), which holds across the cases
and is the contribution we argue most deserves attention beyond research
software engineering. The remaining sections present the cases (§5),
synthesize the cross-axis findings (§6), and name the limitations (§7).

\subsection{5 Case studies}\label{case-studies}

Four case studies cover the axes of §4 individually. Case A (§5.1) is a
solo deep-research lineage that exercises the wiki's depth and
durability under intensive single-author use. Case B (§5.2) is a
multi-author project whose retroactive-audit cascade carries the
strongest single quantitative result in the paper. Case C (§5.3) is the
in-progress multi-AI-agent deployment at Project M, reported as a design
with the load-bearing live evidence pending the Cursor integration. Case
D (§5.4) is a cross-domain educational variant that exercises the
multi-domain axis through a designed-and-scripted tutoring probe.

\subsection{5.1 Case A: Solo deep research with preserved iteration
history}\label{case-a-solo-deep-research-with-preserved-iteration-history}

The foundation case is one researcher using the wiki intensively over an
extended ingest window. Project R is the first author's
retrieval-research project, instantiated from the template; its wiki
contains 48 cross-linked pages with 907 body cross-references and 237
typed-edge frontmatter declarations, built over twenty-two days of
intensive solo ingest. The case is \emph{solo} on the multi-human axis,
\emph{single-agent} (Claude Code), and \emph{single-domain}: a baseline
against which the other case studies turn up additional axes.

Project R's research trajectory, \emph{as filed in the wiki}, is a
sequence of five architectural iterations, each a page or page-cluster:
problem statement, attempt, result, and cross-references to the next
iteration. The eventual best-performing iteration is the fifth; the
third was a strongly-claimed alternative architecture that the fifth
walked back by name. The third iteration's page initially carried a
strong position that an alternative architecture was fundamentally too
rigid to compete on the benchmark. Three weeks of later experiments
contradicted this: the fifth iteration, a modified form of the same
alternative freed from a bolted-on constraint, closed the gap to within
a handful of percentage points of the dense-retrieval baseline. The
original claim was wrong in a specific way, and the wiki absorbed the
correction in the same shape Case B's audit cascade uses: the original
page was not deleted; a new page was filed alongside it with a
cross-reference back, and a \emph{Status} note pointing forward was
added to the original. A reader returning to the project a year from now
encounters the trajectory in place.

A second property surfaces that the other cases do not exercise as
directly: the wiki captures lineage \emph{across projects}. Project R's
core retrieval mechanism extends an antecedent from Project Q (a prior
collaboration at the same research center) which can be viewed as a
single-step instance of the architecture Project R operates over a
larger graph. The wiki captures this connection on the page that defines
Project R's core mechanism, so prior-art context lives in the same
artifact as the project's day-to-day notes. The architectural-scale
preservation of the third iteration (§6.1) and the cross-project lineage
capture both motivate the multi-domain reading of §4.3.

\subsection{5.2 Case B: Multi-author collaboration with quantified
compounding}\label{case-b-multi-author-collaboration-with-quantified-compounding}

The most direct test of the multi-human axis is a two-researcher project
we ran in parallel with the development of the template, here referred
to as Project F (private repository). Project F is an adaptive web
scraper probing the hypothesis that \emph{competency questions can
replace ontologies} as the interface to LLM-based extraction. The wiki
at the center is a common information space shared by two researchers,
R1 and R2, plus their LLM coding agents.

Project F instantiates the pattern \emph{twice}: a \emph{code wiki}
maintained by the developers via their coding agents, and a \emph{memory
wiki} maintained by the running application itself at runtime. The two
share the same schema but have different authors, lifecycles, and
consumers, a small sub-pattern worth naming: the template is reusable at
multiple levels within a single project. In what follows, ``the wiki''
refers to the code wiki where the articulation work between R1 and R2
lives.

\subsubsection{5.2.1 An articulation
moment}\label{an-articulation-moment}

On a Wednesday in late May, R2 contributed a substantial demo branch
(new LLM backend, an orchestrator wrapping the scrape-ask-extract cycle,
a concept registry, an Obsidian-style wiki writer) and filed the
substance in the code wiki as Experiment 11, demo-scoped.

R1, returning the next day, ran end-to-end validation against a
three-entity ground-truth set. The validation surfaced a red flag: the
agent's answers on one entity carried a literal source attribution
reading \emph{``Prior knowledge of {[}entity{]}; visible navigation
items reference products and categories\ldots{}''} The agent was filling
answers from training data when it could not read the page evidence,
directly at odds with the project's working hypothesis. R1 wrote up the
failure in Experiment 11, cross-referenced into the prior experiments it
threatened.

\subsubsection{5.2.2 A retroactive audit, made possible by
provenance}\label{a-retroactive-audit-made-possible-by-provenance}

The ``Prior knowledge'' string pointed to a deeper question: were
earlier experiments' coverage claims also affected? The pipeline's
\texttt{questions\_answered} metric had been treating any model
response, including honest hedges like \emph{``not visible on this
page''}, as a successful answer. The earlier experiments had reported
20-of-20 question coverage; if hedges were being counted, that number
was inflated. Because the wiki preserved per-experiment run summaries
with full per-question detail, R1 could \emph{retroactively audit}
without re-running. A hedge-phrase filter against the saved summaries
produced Table 1:

\begin{longtable}[]{@{}
  >{\raggedright\arraybackslash}p{(\linewidth - 6\tabcolsep) * \real{0.2500}}
  >{\raggedright\arraybackslash}p{(\linewidth - 6\tabcolsep) * \real{0.2500}}
  >{\raggedright\arraybackslash}p{(\linewidth - 6\tabcolsep) * \real{0.2500}}
  >{\raggedright\arraybackslash}p{(\linewidth - 6\tabcolsep) * \real{0.2500}}@{}}
\toprule\noalign{}
\begin{minipage}[b]{\linewidth}\raggedright
Experiment
\end{minipage} & \begin{minipage}[b]{\linewidth}\raggedright
Original claim
\end{minipage} & \begin{minipage}[b]{\linewidth}\raggedright
Audited (evidence-only)
\end{minipage} & \begin{minipage}[b]{\linewidth}\raggedright
Re-run with fix
\end{minipage} \\
\midrule\noalign{}
\endhead
\bottomrule\noalign{}
\endlastfoot
Exp 7 (first manufacturer, deepen) & 20/20 & \textbf{14/20} &
\textbf{18/20} \\
Exp 8 (second manufacturer, cross-entity) & 20/20 & \textbf{12/20} &
\textbf{18/20} \\
\end{longtable}

R1 implemented the fix as a defensive filter at four pipeline points
(merge, plateau computation, work-remaining accounting, deepen-time
sanitization). The fix was committed as a single change
(\texttt{1cf819e}) and triggered a redux of both experiments under the
now-honest metric. The redux produced \textbf{exact parity} between the
two manufacturers at 18-of-20 evidence-based answers, recovering the
cross-entity transfer claim of Experiment 8 at corrected coverage.
Concept extraction tripled on the deepest target (16 → 54 canonical
concepts) because the fix corrected an over-eager plateau stop that had
been cutting page exploration short. Figure 3 shows the temporal
sequence.

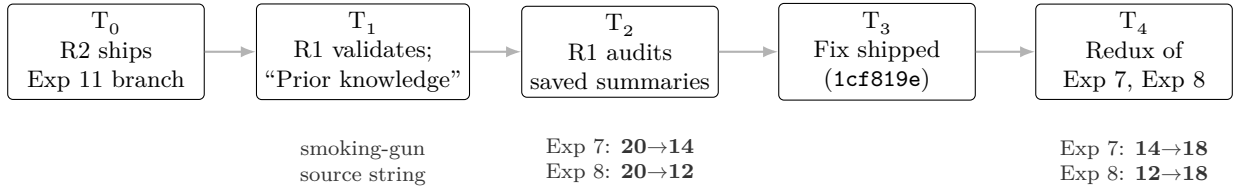
\begin{figure}[h]
\centering
\begin{tikzpicture}[
  event/.style={draw, rectangle, rounded corners=2pt,
                minimum width=2.6cm, minimum height=1cm,
                align=center, font=\footnotesize, inner sep=3pt},
  arrow/.style={-{Latex[length=2mm]}, thick, gray!60},
  annot/.style={font=\scriptsize, align=center, gray!40!black}
]

\node[event] (e1) at (0, 0) {T$_0$\\R2 ships\\Exp~11 branch};
\node[event] (e2) at (3.4, 0) {T$_1$\\R1 validates;\\``Prior knowledge''};
\node[event] (e3) at (6.8, 0) {T$_2$\\R1 audits\\saved summaries};
\node[event] (e4) at (10.2, 0) {T$_3$\\Fix shipped\\(\texttt{1cf819e})};
\node[event] (e5) at (13.6, 0) {T$_4$\\Redux of\\Exp~7, Exp~8};

\draw[arrow] (e1) -- (e2);
\draw[arrow] (e2) -- (e3);
\draw[arrow] (e3) -- (e4);
\draw[arrow] (e4) -- (e5);

\node[annot, below=0.4cm of e2] {smoking-gun\\source string};
\node[annot, below=0.4cm of e3] {Exp~7: \textbf{20$\to$14}\\Exp~8: \textbf{20$\to$12}};
\node[annot, below=0.4cm of e5] {Exp~7: \textbf{14$\to$18}\\Exp~8: \textbf{12$\to$18}};

\end{tikzpicture}
\caption{Case B audit cascade. T$_0$ through T$_4$ are sequential events visible from the wiki's git log, spanning three days. The pipeline's \texttt{questions\_answered} metric had been counting honest model hedges as positive evidence. The original 20-of-20 coverage claims for two prior experiments were revised down to 14 and 12 evidence-based answers at T$_2$, then up to 18 and 18 at T$_4$ after the fix. Both the original inflated claims and the corrected claims remain in the wiki; the trajectory is the documentation.}
\label{fig:audit-cascade}
\end{figure}

The architectural finding, that one generic question set transfers
across heterogeneous manufacturers, survived the audit only because the
wiki preserved enough provenance to \emph{do} the audit in the first
place. Three properties carry into §6: provenance as awareness;
retroactive correction as compounding (the audit produced new pages that
amplified the prior ones rather than replacing them); failure-path
preservation (the original 20-of-20 claims remain alongside the audit
and redux, so a reader of Experiment 7 today learns the trajectory).

\subsection{5.3 Case C: In-progress multi-agent deployment (design
report)}\label{case-c-in-progress-multi-agent-deployment-design-report}

Case C reports on an \emph{in-progress} externally-funded project at the
same host research computing center as Cases A and B, here referred to
as Project M, that intends to use the llm-wiki template as a shared
knowledge backbone across multiple coding assistants. The Claude Code
overlay is live at Project M today, writing log entries that carry
\texttt{by:\ \textless{}human\textgreater{}\ via\ claude-code}; the
Cursor overlay is in integration, and an OpenAI-family overlay is
planned. Live multi-agent attribution examples are therefore not yet
available; we report the architectural design that the overlay system
supports and stage the empirical confirmation for a future revision.

Three properties of the design are not standard in multi-agent systems
that share memory. \emph{Attribution distinguishes
which-human-via-which-agent}: a
\texttt{by:\ A.\ Smith\ via\ claude-code} line next to a
\texttt{by:\ A.\ Smith\ via\ cursor} line reveals the same person
working through two toolchains, while
\texttt{by:\ A.\ Smith\ via\ claude-code} next to
\texttt{by:\ B.\ Jones\ via\ claude-code} reveals two people through one
toolchain; neither view is collapsed. \emph{Policy is agent-agnostic}:
the two policy files (\texttt{discipline-gates.md} and
\texttt{verification-gate.md}) sit at a layer above the overlays and are
referenced from each overlay's per-operation procedure rather than
copied, so a new overlay inherits the same gates the existing ones
enforce. \emph{One-commit-per-log-entry survives multi-agent writes}:
the attribution rule keeps \texttt{git\ blame} on the log file a
faithful per-entry record even when two agents are active in parallel on
the same wiki.

This case is evidence of \emph{design}, not \emph{practice}: the
load-bearing live observation, two heterogeneous coding assistants
producing the right \texttt{by:} lines on the same wiki across the same
day's work, awaits the Cursor integration. We treat the design report as
a \emph{predictive} finding and name the gap in §7.

\subsection{5.4 Case D: Cross-domain variant for undergraduate
education}\label{case-d-cross-domain-variant-for-undergraduate-education}

Case D exercises the multi-domain axis (§4.3) by instantiating the
template through a \emph{variant} rather than the base. The variant,
called the \emph{tutor variant}, supplies a Socratic system prompt
(appended at session launch via Claude Code's
\texttt{-\/-append-system-prompt} flag) that turns the agent into a
Socratic tutor: instead of producing an answer, the agent scaffolds the
student's reasoning and asks a sequence of questions that lead the
student to the answer. The base template's wiki schema, three
operations, agent overlays, security model, and discipline gates (§3)
are retained unchanged. What the variant adds is exactly the things a
base-template user would not want by default: the Socratic prompt, the
launcher that wires it in, and a mode-specific CLAUDE.md section that
warns the agent against delivering answers directly. The placement rule
from §4.3 is at work: pedagogy belongs in the variant, not the base.

The derived project is a wiki for an undergraduate microelectronics lab
module, instantiated from the curriculum source named in the
Acknowledgements. The wiki carries roughly 30 pages organized into three
clusters: \emph{Components}, \emph{Concepts}, and \emph{Code patterns}.
Typed edges encode the prerequisite graph: the \emph{RGB LED} page
declares \texttt{extends:\ LED-Basics},
\texttt{requires:\ Common-Anode-vs-Common-Cathode},
\texttt{uses:\ Pulse-Width-Modulation}. These edges license the agent to
traverse from the component the student is wiring up to the concept the
component depends on, without leaving the wiki.

The case's design probe is a paired-video contrast, scripted
illustrations rather than recordings of observed student behavior. In
Video 1 a scripted student uses a generic LLM to complete Lab 1; the LLM
produces a complete answer in two minutes, the student copies and
submits. In Video 2 the same scripted student starts the same question
against the course wiki under the tutor variant; the agent
soft-redirects from ``give me the code'' to ``let us walk through
\emph{why}'', surfaces the LED-Basics page, walks through Ohm's law, and
watches the student compute the resistor size. A diagnostic question,
\emph{if you had used a blue LED instead of a red one, would 220Ω still
be the right resistor?}, follows in both videos: the Video 1 student
hesitates and guesses; the Video 2 student recomputes from forward
voltage. The contrast is the case's pedagogical hook, attributable in
the design to whether the agent surrounds the student with the course's
own knowledge structure. An empirical study with real students is a
planned follow-up (§7).

\subsection{6 Findings and discussion}\label{findings-and-discussion}

Three properties of the llm-wiki pattern hold across the axes of §4 and
across the four case studies of §5. We lead with the property we argue
most deserves attention beyond research software, then the property that
earns the architecture's trust claims, then the property that makes
cross-domain reach tractable.

\subsubsection{6.1 Failure paths survive}\label{failure-paths-survive}

The least-recognized property of the pattern, and its largest
contribution beyond research-software practice, is that the wiki
preserves \emph{what did not work} alongside what did. Case A retains an
abandoned architectural iteration on the same wiki where the corrected
successor was filed, with cross-references threading the trajectory; a
reader arriving at the third-iteration page is led to the page that
walked back its strongest claim. Case B retains the original inflated
20-of-20 coverage claims next to the audit that exposed them (14 and 12
evidence-based answers) and the redux that recovered them to 18 and 18
(Table 1); the trajectory is the documentation. The template's own wiki
carries a \emph{Lessons-learned} category capturing PR-review failure
modes from its development, including a hook-type confusion that broke
the original PostToolUse hook design until it was diagnosed under live
use.

This addresses a structural blind spot. Journals reward the corrected
result and treat the abandoned path as noise; pre-prints get superseded
by camera-ready versions that delete the qualifications. Concurrent
extensions to the pattern have taken the supersession instinct further
into the artifact itself: \emph{LLM Wiki v2} (§2.3) adds confidence
scoring and explicit supersession of stale claims. We argue the opposite
move. Supersession optimizes for the \emph{current} reader's clarity;
preservation optimizes for the \emph{next} reader's ability to learn
from the trajectory. The mechanism is the append-only convention (§4.1)
combined with the verification gate (§3.3) insisting new claims be filed
alongside, not in place of, the claims they qualify. The discipline
pushes preservation \emph{upstream} of publication into the artifact
researchers work in day-to-day, a different point of intervention from
the metascience response at the publication layer (Ioannidis, 2005).

\subsubsection{6.2 Errors become visible}\label{errors-become-visible}

The cases surface a recurring failure mode: an agent files projections
or cross-corpus comparisons \emph{as if they were measured facts}, and
downstream readers cannot tell which claims are which. Case B's
\texttt{questions\_answered} metric inflation operated by treating model
hedges as positive evidence; this is exactly the failure category the
verification gate's \emph{numerical claims} check (§3.3) names directly,
and the audit was the only thing that prevented the wiki from
accumulating bad numbers. Case A's verification gate operates as the
\emph{internal} check against the same failure mode when no second human
catches projections. The agent-honesty mechanisms (§3.3) do not force
the agent to be honest by construction; they make dishonesty
\emph{visible} and \emph{correctable}, which is a weaker property and,
as Case B demonstrates, an empirically sufficient one. Provenance
reinforces the property: the
\texttt{by:\ \textless{}human\textgreater{}\ via\ \textless{}agent\textgreater{}}
line treats provenance as a \emph{primitive of awareness} (Dourish and
Bellotti, 1992) rather than as bookkeeping, and Case B's audit was
\emph{possible because} per-experiment provenance was preserved with
enough granularity to audit retroactively. Compounding follows: Case A's
wiki (§5.1) and Case B's audit producing new pages that \emph{amplified}
prior pages rather than replacing them are within-case demonstrations of
accumulation. A controlled comparison against alternative substrates
(Notion, plain Git, per-agent built-in memory) is future work (§7).

\subsubsection{6.3 Designs adapt locally}\label{designs-adapt-locally}

The three-tier governance model (§4.3) is a small design move with an
unusual property: it makes \emph{appropriation} legible by carving out a
tier for it. Communities-of-practice work has typically treated
appropriation as an emergent property of use (Lave and Wenger, 1991)
rather than as a built-in tier; the variant template is a counterexample
worth studying. Case D demonstrates that the same base can produce a
research-software wiki (Case A) and an undergraduate-pedagogy wiki
without features leaking: the Socratic prompt, the prerequisite graph,
and the diagnostic-question contrast live in the variant and in the
derived project; the base template's documentation contains no pedagogy.
The mechanism is the placement rule with the \texttt{ALWAYS\_FILES} sync
contract; the graduation trigger promotes variant-level features to the
base when two or more variants converge.

For practitioners: install the discipline gates and verification gate
from day one; when the wiki feels like overhead, lower the bar by filing
\texttt{untyped} placeholder pages rather than skipping (the compounding
claim depends on the \emph{act} of filing); fork into the variant tier
only for groups whose \emph{class} of projects shares a behavior
contract the base cannot provide. The wiki can be browsed via the file
system, GitHub's wiki view, or Obsidian-style clients; the substrate is
the same across surfaces.

\subsection{7 Limitations}\label{limitations}

The four case studies, the three-axis architecture, and the
cross-cutting findings together make claims the evidence does not yet
fully establish. The Claude Code overlay is the only one with end-to-end
behavioral validation; Cursor is structurally validated only, and the
``none'' overlay is untested. Project M (§5.3) is the natural setting
for closing the Cursor gap once its integration ships. Longitudinal
evidence is small: the compounding numbers in §6 come from a single
22-day window of solo writing, and Case B is two researchers rather than
a team. Adversarial prompt injection in untrusted content the agent
reads remains uncaught by the discipline and verification gates, which
target the agent's \emph{own} content mistakes rather than hostile
inputs. The three-tier governance model is well-specified for two or
three variants but increasingly untested at four or five. Most
consequentially, the four-case methodology establishes that the
architecture \emph{works on each axis}; it does not compare against
alternative substrates (Notion, plain Git, per-agent built-in memory). A
controlled study against any of these on a measurable outcome
(time-to-rediscovery, retroactive-correction rate, cross-session
knowledge retention) is the natural next step. The
three-axes-simultaneous case is named here as a near-term opportunity
rather than a current gap.

\subsection{8 Conclusion}\label{conclusion}

The llm-wiki pattern was first proposed as a solution to a per-session
memory problem. We have argued that, when instantiated through
\texttt{llm-wiki-memory-template}, it supports heterogeneous
collaborative knowledge work along three distinct axes (§4), with case
studies exercising the axes one and two at a time. The central finding
is the cross-axis property of \emph{failure-path preservation} (§6.1):
abandoned approaches, walked-back claims, and audited-down numbers
survive alongside the eventual best result, and the trajectory is the
documentation. We argue this is a contribution to research practice as
much as to research-software infrastructure.

The variant population is growing. Beyond the educational variant
reported here, codebase-onboarding and grant-authoring variants are
under design discussion; the Governance page in the base template's own
wiki names three open questions whose resolution becomes urgent as the
variant population grows: variant discovery (registry, tag convention,
word-of-mouth?), versioning (pin to SHA or rebase to base HEAD?), and
cross-variant features (when does a feature graduate from a variant into
the base?). The template is offered as a copyable starting point for any
research, educational, or workforce-development team whose LLM agents
need to remember and whose human collaborators need a place where memory
belongs. It is at
https://github.com/crcresearch/llm-wiki-memory-template. Adoptions,
variants, and reports of failure modes we did not anticipate are
welcome.

\subsection{Acknowledgements}\label{acknowledgements}

We thank Chris Frederick for the web-scraping project collaboration that
produced Case B's audit cascade, and Charles Vardeman, whose internal
discussions on wiki-as-memory at the host research computing center
predated this template and substantially shaped the team's thinking
before either Karpathy's blog-form proposal or tonbi's working template
existed. We thank the staff and leadership of the Center for Research
Computing at the University of Notre Dame for institutional support, and
the engineers and researchers on Project M. We acknowledge A. Karpathy
for proposing the llm-wiki pattern and tonbi for the working template
implementation that demonstrated it. Curriculum content for Case D is
adapted from revision a (2026) of the Purdue SCALE program's
\emph{Introduction to Engineering with Microelectronics} module, led by
M. Riley.

\subsection{References}\label{references}

Bannon, L., \& Bødker, S. (1997). Constructing common information
spaces. In \emph{Proceedings of the European Conference on
Computer-Supported Cooperative Work (ECSCW'97)}, 81--96. Kluwer Academic
Publishers.

CRC Research. (n.d.). llm-wiki-memory-template. GitHub repository.
\url{https://github.com/crcresearch/llm-wiki-memory-template}. Accessed
2026-05-26.

Dourish, P. (2001). \emph{Where the Action Is: The Foundations of
Embodied Interaction}. MIT Press, Cambridge, MA.

Dourish, P., \& Bellotti, V. (1992). Awareness and coordination in
shared workspaces. In \emph{Proceedings of the ACM Conference on
Computer-Supported Cooperative Work (CSCW'92)}, 107--114. ACM.

Edge, D., Trinh, H., Cheng, N., Bradley, J., Chao, A., Mody, A., Truitt,
S., \& Larson, J. (2024). From local to global: A graph RAG approach to
query-focused summarization. \emph{arXiv preprint} arXiv:2404.16130.
(GraphRAG)

Ghelbur, E. (2026, April 29). I rebuilt Karpathy's LLM Wiki: Here's
what's missing from the original. \emph{The AI Operator}.
\url{https://theaioperator.io/p/i-rebuilt-karpathys-llm-wiki-heres}.
Open-source implementation at
\url{https://github.com/eugeniughelbur/obsidian-second-brain}. Accessed
2026-05-26.

Gutiérrez, B. J., Shu, Y., Gu, Y., Yasunaga, M., \& Su, Y. (2024).
HippoRAG: Neurobiologically inspired long-term memory for large language
models. In \emph{Advances in Neural Information Processing Systems
(NeurIPS'24)}.

Gutiérrez, B. J., Shu, Y., Qi, W., Zhou, S., \& Su, Y. (2025). HippoRAG
2: From RAG to memory, non-parametric continual learning for large
language models. In \emph{Proceedings of the International Conference on
Machine Learning (ICML'25)}.

Ioannidis, J. P. A. (2005). Why most published research findings are
false. \emph{PLoS Medicine}, 2(8), e124.

Karpathy, A. (2026, April). LLM wiki. GitHub gist.
\url{https://gist.github.com/karpathy/442a6bf555914893e9891c11519de94f}.
Announced on X (formerly Twitter),
\url{https://x.com/karpathy/status/2039805659525644595}. Accessed
2026-05-23.

Lave, J., \& Wenger, E. (1991). \emph{Situated Learning: Legitimate
Peripheral Participation}. Cambridge University Press, Cambridge.

Lewis, P., Perez, E., Piktus, A., Petroni, F., Karpukhin, V., Goyal, N.,
Küttler, H., Lewis, M., Yih, W., Rocktäschel, T., Riedel, S., \& Kiela,
D. (2020). Retrieval-augmented generation for knowledge-intensive NLP
tasks. In \emph{Advances in Neural Information Processing Systems
(NeurIPS'20)}.

Packer, C., Wooders, S., Lin, K., Fang, V., Patil, S. G., Stoica, I., \&
Gonzalez, J. E. (2023). MemGPT: Towards LLMs as operating systems.
\emph{arXiv preprint} arXiv:2310.08560.

redmizt. (2026, April). Beyond the Wiki: Scaling Karpathy's LLM Wiki
Pattern for Multi-Agent Production. GitHub gist.
\url{https://gist.github.com/redmizt/968165ae7f1a408b0e60af02d68b90b6}.
Accessed 2026-05-26; the gist no longer resolves at this URL as of the
access date, but its title and abstract are indexed by major search
engines.

rohitg00. (2026, April 6). LLM Wiki v2: Extending Karpathy's LLM Wiki
pattern with lessons from building agentmemory. GitHub gist.
\url{https://gist.github.com/rohitg00/2067ab416f7bbe447c1977edaaa681e2}.
Accessed 2026-05-26.

Schmidt, K., \& Bannon, L. (1992). Taking CSCW seriously: Supporting
articulation work. \emph{Computer Supported Cooperative Work (CSCW): An
International Journal}, 1(1), 7--40.

Star, S. L., \& Bowker, G. C. (1999). \emph{Sorting Things Out:
Classification and Its Consequences}. MIT Press, Cambridge, MA.

Strauss, A. (1985). Work and the division of labor. \emph{The
Sociological Quarterly}, 26(1), 1--19.

Strauss, A. (1988). The articulation of project work: An organizational
process. \emph{The Sociological Quarterly}, 29(2), 163--178.

tonbi {[}tonbistudio{]}. (2026, April 8). llm-wiki: Open-source template
for building LLM-powered knowledge bases following Karpathy's LLM Wiki
pattern. GitHub repository.
\url{https://github.com/tonbistudio/llm-wiki}. Accessed 2026-05-26.

\end{document}